# Age Estimation Based on Graph Convolutional Networks and Multi-head Attention Mechanisms

Miaomiao Yang[#]
Xi'an Jiaotong-liverpool University
Jiangsu, China

Changwei Yao[*,#]
Southern University of Science and Technology
Shenzhen, China
yaocw@mail.sustech.edu.cn

Shijin Yan[#]
Southwest Jiaotong University
Sichuan, China

*Abstract*—Age estimation technology is a part of facial recognition and has been applied to identity authentication. This technology achieves the development and application of a juvenile anti-addiction system by authenticating users in the game. Convolutional Neural Network (CNN) and Transformer algorithms are widely used in this application scenario. However, these two models cannot flexibly extract and model features of faces with irregular shapes, and they are ineffective in capturing key information. Furthermore, the above methods will contain a lot of background information while extracting features, which will interfere with the model. In consequence, it is easy to extract redundant information from images. In this paper, a new modeling idea is proposed to solve this problem, which can flexibly model irregular objects. The Graph Convolutional Network (GCN) is used to extract features from irregular face images effectively, and multi-head attention mechanisms are added to avoid redundant features and capture key region information in the image. This model can effectively improve the accuracy of age estimation and reduce the MAE error value to about 3.64, which is better than the effect of today's age estimation model, to improve the accuracy of face recognition and identity authentication.

## I. INTRODUCTION

### A. Motivation

Age estimation is a method of predicting an individual's age by extracting information from facial images. An existing survey in such a field generally focuses on inferring the age range or specific age of individuals, concerning the disciplines of computer vision and machine learning. Considerable attention has been drawn to age estimation in the field of artificial intelligence, both domestically and internationally. On the one hand, the accuracy of age prediction has not reached an ideal level due to the influence of numerous complex factors, such as lifestyle, identities, gender, race, and facial expressions of individuals, which causes it challenging to ensure consistency for non-essential feature information within the same age group [1]. On the other hand, age estimation holds significant practical significance and relevant applications are risk identification and security measurement. For instance, age detection can prevent the sale of cigarettes and alcohol to minors, and facial image analysis in video surveillance can help approximate the age range of a suspect. Additionally, age detection can enable businesses and governments to conduct population demographics and market research in a more effective way [2].

Automatic age estimation involves a series of processes, involving image data preprocessing, feature extraction, model training, and age value prediction. Feature extraction, a significant component of automation, is typically accomplished with deep learning techniques, which have demonstrated their effectiveness in learning high-level features from images. Among the various methods utilized in age estimation, Convolutional Neural Networks (CNNs) and Transformer models have gained significant popularity. CNN, initially proposed by LeCun [3], has been widely applied in signal processing, image classification, and image recognition due to its efficiency in extracting local information from images using Convolutional kernels. Transformer models, based on the optimization of Recurrent Neural Networks(RNN), excel in operating long sequences and can conduct efficient para computation during both training and inference stages [4]. While both frameworks have achieved success in image recognition, they still exhibit limitations and flaws when applied to age prediction tasks. Notably, CNN models are sensitive to the variation of factors such as lighting, pose, and shadows during image processing, significantly affecting the accuracy of age estimation. Conversely, Transformer models, unlike CNN, do not explicitly consider the spatial structure information of images during the feature learning process, resulting in potential challenges in capturing fine details such as wrinkles in facial images [5].

In the context of age estimation, it is imperative to minimize the influence of extraneous factors and environmental variations in the image, while also incorporating spatial information from local regions. This paper proposes a novel approach that combines the RCN neural network framework with a multi-head attention mechanism to enhance the learning process. The proposed model utilizes Graph Convolutional Networks (GCN) to capture relationships and contextual information among faces. By simultaneously extracting local and global features through multiple stacked layers, the model aims to address the limitations of CNN in capturing complex inter-node relationships, which often lead to oversensitivity to interfering factors. Additionally, the multi-head attention mechanism is employed to identify the most informative features, focusing on the aspects that truly contribute to the accurate estimation of age and reducing the interference caused by irrelevant image information.

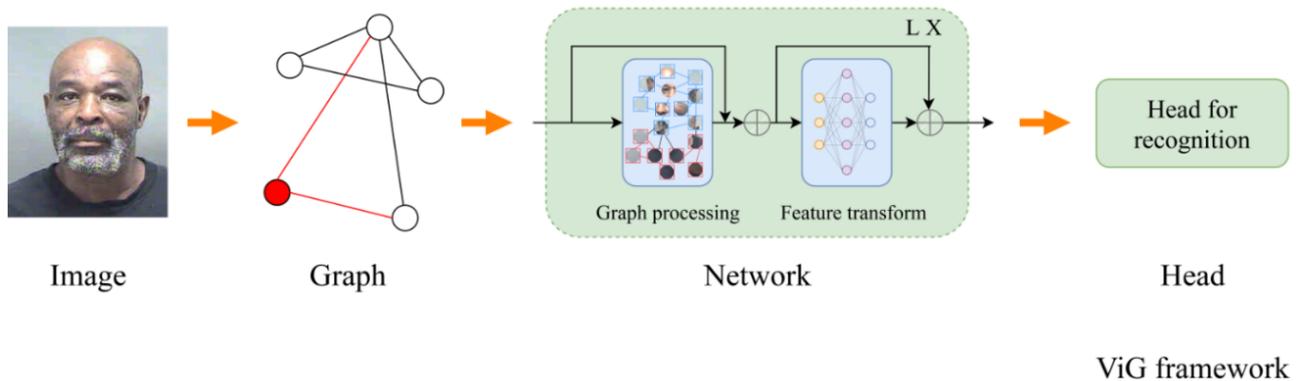

Fig. 1 ViG framework

*B. Our Contributions*

- Distinguished from prevailing data modeling paradigms, our contribution introduces an innovative fusion model. This model seamlessly integrates the GCN as its foundational architecture while harnessing a multi-head attention mechanism to enhance the efficiency of the feature extraction learning process. By tapping into the inherent capabilities of GCN, the model endeavors to capture significant local region information intricately embedded within images. This concurrent accommodation of contextual associations among discrete facial images lends to a holistic synthesis. This harmonious integration manifests in the palpable amelioration of errors inherent in the age prediction process.

- In a notable departure from the customary recourse to Euclidean distances for computational purposes, our proposed model engenders a paradigmatic shift. This departure is characterized by the utilization of non-Euclidean distances within the intricate fabric of the GCN neural network. This innovative approach augments the depth and nuance of our comprehension, casting a comprehensive light upon the intricate relationships that constitute the foundational scaffold of facial image data.

## II. RELATED WORK

Age estimation through facial image analysis is a crucial and extensive academic topic in the field of artificial intelligence. It aims to automatically determine the precise age of an individual by leveraging machine learning models to extract meaningful features from facial images. This technology finds diverse applications in domains such as video surveillance, population statistics, and demographic analysis. However, the existing study has not achieved a high level of accuracy in age estimation. Consequently, the current research trend revolves around developing novel approaches that involve the construction of different learning and auxiliary models. These models aim to mitigate the influence of irrelevant information and capture the intrinsic correlations among facial images at a global level.

In recent years, numerous studies have proposed various fusion models combining neural networks with auxiliary algorithms or making modifications to existing network architectures, aiming to capture more valuable local features from facial images and improve the accuracy of output. Chang et al. introduced a CNN2ELM model that utilizes CNN for feature extraction and ELM for individual age estimation, harnessing the strengths of convolutional neural networks and deep learning machines. Sun et al. [6] proposed a Conditional Distribution Learning(DCDL) method based on a mathematical formula for distribution learning, which reflects the adaptivity when learning essential age-related features in the feature extraction process, assigning higher weights to influential age-related features. Liu et al. [7] extended the CNN model by introducing a similarity-aware deep adversarial learning method, utilizing age difference information in the synthesized feature space for robust age estimation. This approach ensures smoother prediction by considering global image features [8]. These studies aim to reduce the sensitivity of individual CNN frameworks to irrelevant variations in local image regions and explore the fusion of multiple CNNs to capture spatial information in images [9]. However, the ability to effectively capture the interconnections among global facial images has not achieved the requirement of practical application, which remains a crucial challenge in current facial age estimation research [10].

In age estimation from facial images [11], it is essential to capture the global information of facial images while considering their spatial structure and established relationships among information nodes. GCN were utilized to extract feature information and learn the connections between facial data and age. To mitigate interference from irrelevant information, such as background, expression, race, and gender in facial images, and to capture the most valuable facial information, a multi-head attention mechanism was integrated to optimize the age estimation results. This approach enables a more effective way to conduct non-Euclidean distance-type graph-structured information in facial images, taking into account both global and local information.

## III. PRELIMINARY INFORMATION

*Self-Attention Mechanism*

In this section, we present the structure of the Multi-Head Attention mechanism. This mechanism involves mapping and calculating attention on the input sequence multiple times, producing distinct attention subspaces with independent parameters. The strategy achieves efficiency by performing

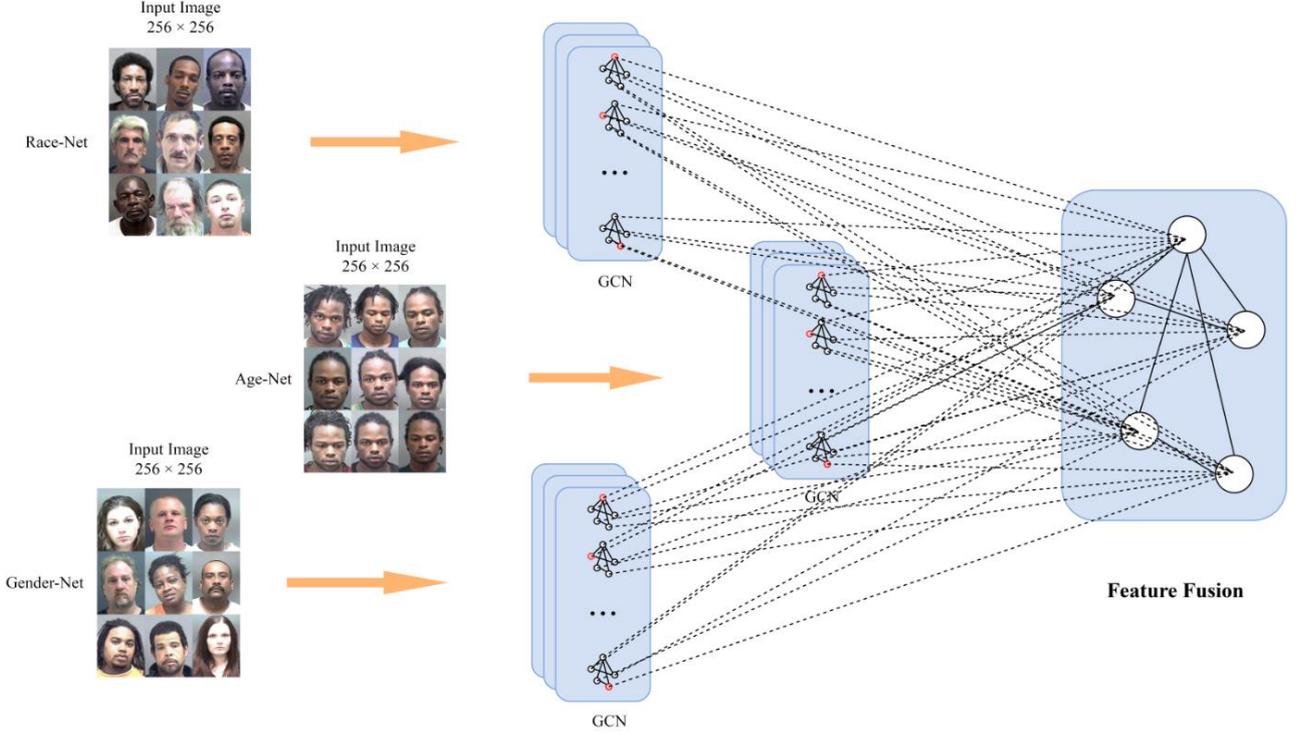

Fig. 2 The feature fusion of GCN

parallel attention computations and capturing more semantic information.

We postulate the query as $Q \in R^{T_Q \times d_Q}$, the key as $K \in R^{T_K \times d_K}$ and the value $V \in R^{T_V \times d_V}$. Here $d_Q$, $d_K$ and $d_V$ are the feature dimensions of query, key and value, $T_Q$, $T_K$ and $T_V$ denote the sequence length of query, key and value respectively.

Within each distinct attention subspace, the query vectors engage in the computation of attention weights by evaluating their similarity with key vectors. Subsequently, these computed weights are employed to effectuate a weighted summation of the value vectors, culminating in the derivation of the ultimate representation.

The initial step involves the linear projection of the Q, K, and V matrices 't' times, each instance involving a distinct linear projection. Here, $'t'$ signifies the number of attention heads:

$$\hat{Q} = \text{Concat}(QW_1^Q, \dots, QW_i^Q, \dots, Q_m^Q) \quad (1)$$

$$\hat{K} = Concat(KW_1^K, \dots, KW_i^K, \dots, K_m^K) \quad (2)$$

$$\hat{V} = Concat(VW_1^V, \dots, VW_i^V, \dots, V_m^V) \quad (3)$$

where, $W_i^Q \in Rd_Q \times d_m$, $W_i^K \in Rd_K \times d_m$ and $W_i^V \in Rd_V \times d_m$ are learnable parameters and $d_m$ denotes the output feature dimensions. For each head, we then perform the dot-product attention in parallel. Finally, the outputs of all heads will be concatenated together as final values $H_{head}$ as follow.

$$head_i = softmax\left((QW_i^Q)(KW_i^K)^T\right)(VW_i^V) \quad (4)$$

$$H_{head} = Concat(head_1, \dots, head_t) \quad (5)$$

## IV. METHODOLOGY

Within this section, we undertake the formalization of the predicament at hand and expound upon our novel approach comprehensively. Fig. 1 and Fig. 2 serve to visually depict the holistic architecture of our model, delineated into two pivotal components:

Transformation of Images into Graph Representation: The foremost element pertains to the conversion of images into graph-based representations.

Feature Extraction via GCNs employing Multi-Head Strategy: The second facet involves the extraction of pertinent features through the utilization of Graph Convolutional Networks, enriched through the incorporation of a multi-head strategy.

Fig. 1 and Fig. 2 collectively encapsulate the overarching framework of our model, encompassing these essential components.

### A. Problem Definition

In the context of a provided dataset comprising facial images accompanied by corresponding age labels, the primary goal involves the training of a model to autonomously acquire salient features from the images and subsequently allocate each individual to the suitable age group. This undertaking typifies a quintessential supervised learning endeavor, which holds profound ramifications within the domain of computer vision.

### B. Graph Representation Establishment

Vertex: For each image $(H \times W \times 3)$, We divide it evenly into $N(k \times k)$ patches in a grid-like manner, where k is a manually specified hyperparameter. We use a feature vector $x_i \in R^D$ to represent a patch, where $D$ is the feature dimension. So we have $X = [x_1, x_2, ..., x_N]$ responding a set of nodes $V = [v_1, v_2, ..., v_N]$.

Edge: In similarity-based graphs, the presence of edges between two vertices is determined by their degree of similarity. For instance, if we hypothesize that all two different vertices have strong connections, then a fully connected graph would entail $O(N^2)$ edges, necessitating expensive computational resources. To address this issue, a more practical approach involves employing the K-Nearest Neighbor (K-NN) to identify the k closest vertices to each vertex and establish edges based on their proximity, indicating the existence of a connection.

Edge Weights: The edge weights are not simply set to one after K-NN processing. Instead, we treat the edge weights as learnable parameters and incorporate them as additional conditional inputs to the network. This approach allows the model to dynamically adjust the edge weights based on the context of the data.

Feature Transformation: With vertex information X and edge information E, we can easily establish the graph $G(X, E)$. A graph convolutional layer enables information exchange between nodes by aggregating features from their neighbors. Specifically, we use a two-step graph convolutional process. In the first step, it computes the representation for each node by aggregating information from its neighborhood using the relation-specific transformation.

$$h_i^{(1)} = \sigma \left\{ \sum_{r \in R} \sum_{j \in N_i^r} \left( \frac{\alpha_{ij}}{c_{i,j}} W_r^{(1)} g_j + \alpha_{ii} W_0^{(1)} g_i \right) \right\} \quad (6)$$

where, $h_i^{(1)}$ denotes new features for vertex $v_i$, $\alpha_{ij}$ and $\alpha_{ii}$ are the edge weights, $N_i^r$ represents a neighborhood set of $v_i$ within relation $r \in R$. $c_{i,j}$ is the preset normalization constant, $\sigma$ denotes any activation function such as ReLU, $W_r^{(1)}$ and $W_0^1$ both are learnable parameters in the process. In the second step, we utilize the first-step outputs as the inputs to do the same operation.

$$h_i^{(2)} = \sigma \left( \sum_{j \in N_i^r} W_r^{(2)} h_j^{(1)} + W_0^{(2)} h_i^{(1)} \right) \quad (7)$$

where, $W^{(2)}$ and $W_0^{(2)}$ are parameters of these transformation and $\sigma$ is the activation function. Finally, the aggregated features will be divided into $t$ parts, i.e. $head^1$, $head^2$, ..., $head^t$ as the multi-head strategy of graph convolution inputs. These heads are updated with different weights separately and will be concatenated as the final value.

$$h = [head^1 W_{update}^1, head^2 W_{update}^2, ..., head^t W_{update}^t] \quad (8)$$

*C. Age Estimation with GCN*

Network Architecture: Unlike the parallel structure of the Transformer, our approach still adopts the commonly used pyramid architecture of CNNs (i.e. ResNet). Although Transformer has been applied to image processing and has many improved methods, empirical evidence shows that the pyramid structure is more effective. The specific model architecture is shown in Fig. 3.

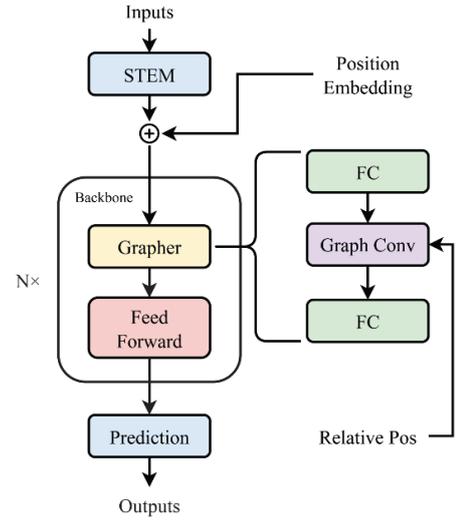

Fig. 1 Our model architecture

Visual Word Embedding: First, the images are transformed into visual word embeddings by passing through multiple CNN layers in the STEM module. Then, the image features are combined with position embeddings to capture extra positional context information. The concatenated result is used as the input to the backbone block.

$$O_{stem} = STEM(X) + PosEmb \quad (9)$$

Backbone Block: In contrast to the previous approach, we have introduced an innovative technique in the backbone architecture. To preserve feature diversity and overcome the over-smoothing issue encountered in deep graph neural network training, we have strategically integrated fully connected layers before and after the graph neural network module. This novel inclusion serves to mitigate feature loss and maintain a higher level of model performance. The Grapher module's processing procedure is characterized as follows:

$$O_{Grapher} = \sigma(GCLayer(O_{stem} W_{in})) W_{out} + O_{stem} \quad (10)$$

To further learn higher-level feature representations from the features and alleviate over-smoothing effects, the output of the Grapher module is subjected to deep processing through two fully connected layers. By incorporating these deep fully connected layers, the model gains the capacity to handle intricate features, thus improving its overall performance and generalization capability.

$$O_{FFN} = \sigma(O_{Grapher} W_1) W_2 + O_{Grapher} \quad (11)$$

The classifier consists of a straightforward stack of two convolutional layers. The final features extracted from the previous layers are passed through this classifier to obtain age estimation.

V. EXPERIMENTS

*A. Benchmark Dataset Used*

MORPH, a widely recognized dataset in the field of computer vision, plays a pivotal role in the advancement of facial image research, encompassing the development of facial recognition, facial attribute analysis, and age estimation algorithms. The dataset involves a substantial collection of face images, exhibiting variations in age, gender, and ethnicity, comprising a total of 55,134 images featuring 13,618 subjects.

The age range of the subjects spans from 16 to 77 years old with an average age of 33. The distribution of images across different age groups within the dataset is presented in Fig. 4. This dataset includes diverse shooting conditions, including variations in lighting environments, people's postures, and facial expressions. Moreover, it contains images capturing individuals at different age periods, supplying researchers with a valuable resource for algorithm testing and training purposes.

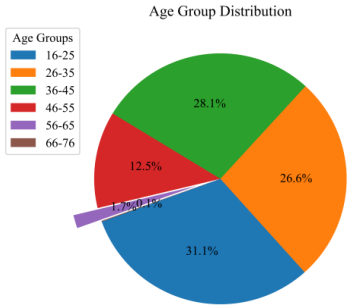

Fig. 2 Statistics for different ages in the dataset

### B. Evaluation Metrics

The employed loss function in this research endeavor is the Mean Absolute Error (MAE), a prevalent metric employed in regression model assessment to quantify the average discrepancy between predicted and actual values. In this specific project, MAE is leveraged to compute the disparity between the estimated age and the ground truth age, thereby serving as a benchmark for evaluating the predictive accuracy of the algorithm. Mathematically, MAE is defined as follows:

$$MAE = \frac{1}{m}\sum_{i=1}^{m}|y_i - \hat{y}_i| \quad (12)$$

where, m in the formula represents the number of samples, which corresponds to the 55,134 images included in the dataset. $y_i$ is the label of each sample, which is the actual age of the sample image. $\hat{y}_i$ denotes the predicted age utilizing the algorithm.

A smaller Mean Absolute Error (MAE) value indicates a declining average discrepancy between the predicted results of the model and the actual values, thus implying a superior fitting capability of the model. By aggregating the individual errors, MAE motivates the amplification of larger errors, enabling researchers to discern their impact on the overall outcome.

### C. Baseline Models

CSOHR: Chang et al. developed a method that leverages the relative-order information among age ranks for age prediction. However, their method is limited to age estimation from faces with nearly neutral expressions.

Bogazici: Gurpinar et al. proposed a two-level system for age estimation, with the use of kernel Extreme Learning Machines (ELM) to classify samples into overlapping age groups and local regressors to estimate a parent age. Its CNNs structure makes it robust to common difficulties in image processing such as pose and illumination as well as occlusions. However, the within-group variance also increases with age, suggesting that this method performs poorly for older samples.

OHRank: Chang et al. proposed the OHRank algorithm to estimate age through images. The principle of this algorithm is based on the relative sequence information of the data set, using the cost-sensitive feature to get a better hyperplane. However, because this algorithm is to develop multiple binary classification framework models for classification tasks, its cost is relatively large, the efficiency is relatively low, and the final result cannot be better improved.

rKCCA: Parkhi et al. proposed to use large-scale data sets (2.6M images, over 2.6K people to traverse deep network training and the complexity of faces to get better methods and programs to improve the algorithm to get better results. However, this model based on large-scale data sets is difficult to reproduce and cannot be applied to small-scale data sets for training.

## VI. RESULT AND DISCUSSION

We compare the performance of our proposed architecture with some baseline models for estimating age based on faces. For the baseline models, a portion was carefully implemented by the original description, while others were directly cropped from the papers. We summarize the performance of these methods on the MORPH dataset in Table 1. To enhance the persuasiveness of the results, the process was repeated ten times, and the average outcomes were adopted as the ultimate representation of age estimation performance. Compared with the baseline methods, it is obvious that our framework achieves better performance, with a lower MAE value, showing the effectiveness of our strategy.

TABLE 1 PERFORMANCE OF METHODS ON THE MORPH

| Model | MAE |
| --- | --- |
| rKCCA | 3.98 |
| CSOHR | 3.74 |
| DeepRank | 3.57 |
| DeepRank+ | 3.49 |
| GCN with Multi-head Attention | **3.47** |

We visually analyze the performance of the proposed GCN model with a multi-head attention mechanism on the training set and validation set, as shown in Fig. 5. According to the images, it can be seen that the MAE of the model decreases steadily on the training set, and fluctuates greatly on the validation set but remains flat. Both eventually converge and eventually stabilize.

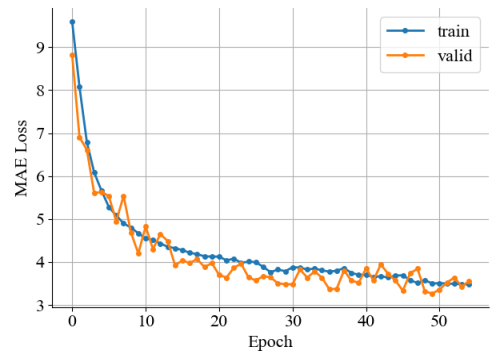

Fig. 3 The MAE of the proposed GCN model on the training set and validation set

Fig. 6 showcases the outcomes of age estimation on the validation set. The visualization demonstrates the commendable performance of the GCN architecture devised in this study across a spectrum of variables, encompassing diverse shooting poses, a range of facial expressions, and variations in occlusion sizes.

Within this representation, a striking observation unfolds: the estimated error for samples that manifest superior performance can plunge below 0.2 years, underscoring the precision of the predictions. Meanwhile, even for samples that exhibit a relatively lower performance level, the error remains confined to less than 1.4 years, signifying the model's resilience in challenging scenarios. Notably, Fig. 6 further attests to the inherent stability of the GCN's features, adeptly surmounting common obstacles such as blurriness, changes in pose, and occlusions.

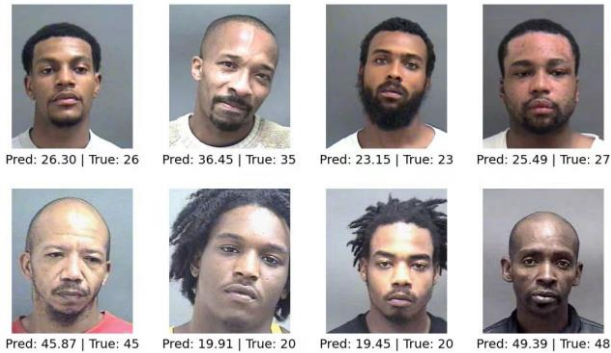

Fig. 4 The estimation results of the validation set

Within Fig. 7, we showcase select instances where our age estimation system encounters setbacks due to various potential factors, including lighting conditions, ethnic variations, and background complexities, particularly noticeable when dealing with the elderly. One of the key reasons for this is the small sample size of older people in the MORPH dataset. According to the statistics in Fig. 4, the sample number of images over 45 years old only accounts for about 13% of the total number. The problem remains to be solved.

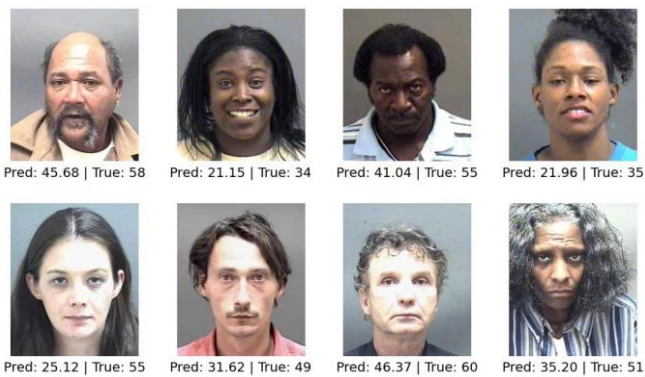

Fig. 5 The age estimation system encounters setbacks

## VII. CONCLUSION AND FUTURE WORK

In this paper, we propose a new model for face age estimation using GCN and multi-head attention mechanisms. Based on GCN, the model can effectively process irregular face images and extract effective image features to achieve high accuracy of age estimation. Furthermore, the multi-head attention mechanism positively influences the model to capture key features, reducing the interference of redundant information and improving the model's accuracy.

Compared with the other existing age estimation algorithms, this model has a significant improvement and reduces the error between estimation and actual age. However, the disadvantage of this model is that the convergence rate is slow, which leads to a longer training time and higher equipment requirements. Therefore, we will use contrast learning to improve graph neural networks in the future to improve model efficiency and reduce training speed. In addition, the problem of prediction bias under different lights and in the elderly is also a direction that needs to be solved, and the data set needs to be further expanded.